\pdfoutput=1
\documentclass[11pt]{article}

\usepackage{acl}
\usepackage{times}
\usepackage{latexsym}
\usepackage{graphicx}
\usepackage{amsmath}
\usepackage[T1]{fontenc}
\usepackage[utf8]{inputenc}

\usepackage{microtype}

\usepackage{inconsolata}

\usepackage{xcolor}
\definecolor{myp}{RGB}{128, 0, 128}
\title{Language Models ``Grok'' to Copy}

\author{Ang Lv$^{1,2}$,\quad Ruobing Xie$^{2}$\footnotemark[1], \quad Xingwu Sun$^{2}$,\quad Zhanhui Kang$^{2}$,\quad\textbf{Rui Yan}$^{1,3}$\thanks{\ \ Corresponding authors: Ruobing Xie (ruobingxie@tencent.com) and Rui Yan (ruiyan@ruc.edu.cn)} \\
  $^1$Gaoling School of Artificial Intelligence, Renmin University of China\\
    $^2$Machine Learning Platform Department, Tencent\\
    $^3$School of Computer Science, Wuhan University\\
  \texttt{\{anglv, ruiyan\}@ruc.edu.cn} \quad  \texttt{ruobingxie@tencent.com} \\
}

\begin{document}
\maketitle
\begin{abstract}
We examine the pre-training dynamics of language models, focusing on their ability to copy text from preceding context—a fundamental skill for various LLM applications, including in-context learning (ICL) and retrieval-augmented generation (RAG).
We propose a novel perspective that Transformer-based language models develop copying abilities similarly to grokking, which refers to sudden generalization on test set long after the model fit to the training set.
Our experiments yield three arguments: (1) The pre-training loss decreases rapidly, while the context copying ability of models initially lags and then abruptly saturates. 
(2) The speed of developing copying ability is independent of the number of tokens trained, similarly to how grokking speed is unaffected by dataset size as long as the data distribution is preserved. 
(3) Induction heads, the attention heads responsible for copying, form from shallow to deep layers during training, mirroring the development of circuits in deeper layers during grokking.
We contend that the connection between grokking and context copying can provide valuable insights for more effective language model training, ultimately improving in-context performance.
For example, we demonstrated that techniques that enhance grokking, such as regularization, either accelerate or enhance the development of context copying. 
\end{abstract}

\section{Introduction}
\label{sec:intro}
Large language models (LLMs) can learn, retrieve, and reason from input context, facilitating various applications such as in-context learning (ICL,~\citealp{brown2020languagemodelsfewshotlearners}) and retrieval-augmented generation (RAG,~\citealp{lewis2020retrieval}).
Despite these achievements, several shortcomings have been reported regarding LLMs' in-context capacities.
For instance, the order of ICL demonstrations matters~\cite{lu-etal-2022-fantastically} and LLMs' awareness of different contextual positions fluctuates~\cite{liu2023lost}.
We believe that studying the mechanisms behind the development of in-context capabilities during pre-training offers valuable insights for enhancing LLMs from a novel perspective.

In this paper, we examine the pre-training dynamics of language models, focusing specifically on their \textbf{context copying} capabilities. 
These capabilities are crucial for various LLM applications, including ICL and RAG.
For example, \citet{olsson2022context} interpret ICL as a process that entails copying and then fuzzy pattern completion. 
Similarly, RAG exhibits this characteristic, as it requires the in-context retrieval of key information, which is then copied (or integrated with additional paraphrasing and reasoning) as the output.
This paper presents empirical evidence demonstrating that Transformer-based language models~\cite{transformer} develop context copying capabilities in a manner akin to ``\textbf{grokking}''~\cite{power2022grokkinggeneralizationoverfittingsmall}. 
Grokking refers to the abrupt improvement in test set generalization long after models have overfit. 

Our experimental method is summarized as follows:
We trained 12-layer Llama models~\cite{touvron2023llamaopenefficientfoundation} using 40 billion tokens and saved checkpoints at regular intervals. 
To evaluate context copying, we presented the models with an input context comprising multiple random token subsequences, each beginning with a unique prefix, and let them complete one of the prefixes presented in the context.
The accuracy of these completions served as a measure of the models' context copying abilities. 
By analyzing the evolution of context copying accuracy and the development of \emph{circuits} (i.e., the subnetworks responsible for completing the specific task) across the saved checkpoints, we argue there is a potential connection between grokking and the development of context copying capabilities, as outlined in the following arguments:

\paragraph{Argument 1: Grokked Context Copying.} We observe that context copying accuracy shows a sudden increase long after the training loss stabilizes, akin to ``grokking'' on the test set when neural networks trained on small training sets.

\paragraph{Argument 2: Token-Count-Independent Grokking Speed.} 

We adjust the batch size to manage the number of tokens trained at specific update steps. 
Results indicate that context copying is developed after certain updates, rather than after processing a specific quantity of tokens. 
Similarly, the data-amount-independent (i.e., token-count-independent) generalization speed is a characteristic of grokking~\cite{wang2024grokkedtransformersimplicitreasoners}.

We found that a higher learning rate speeds up grokked copying, suggesting it occurs at a specific optimization intensity, determined by the learning rate and update steps. 
These experiments underscore the importance of careful hyperparameter selection in training language models for capacities like context copying, as their development isn't necessarily reflected in pre-training loss reduction.

\paragraph{Argument 3: Deeper Circuit Formation.} 
We note that \emph{induction heads}~\cite{olsson2022context}, attention heads responsible for copying tokens, form from shallow to deep layers during training, consistent with research showing deeper circuits form in Transformers after grokking~\cite{wang2024grokkedtransformersimplicitreasoners}.

Based on the novel perspective that language models grok to copy, we pre-trained language models using regularization techniques, which are known to enhance grokking.
These techniques lead to either faster copying acquisition or higher accuracy.
Our findings highlight a promising and efficient research approach: developing improved language models with enhanced in-context performance by leveraging an understanding of grokking. 
This efficiency arises from the fact that studies on grokking can utilize smaller, synthesized datasets, thereby avoiding the extensive and resource-intensive trials required for directly pre-training language models.

\section{General Setup}
\label{sec:setup}

\begin{figure}[t]
% \vspace{-2mm}
    \centering
    \includegraphics[width=\linewidth]{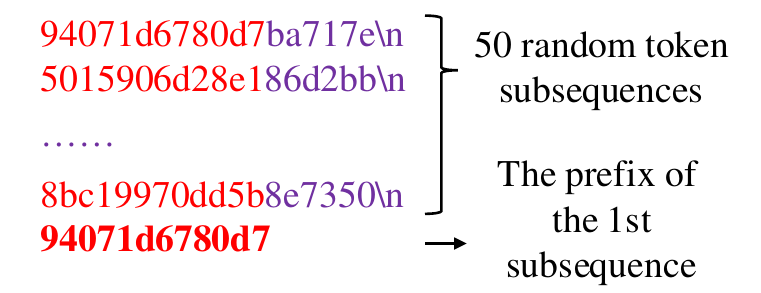}
    \caption{An test input example when $i=1$. The correct completion of this input should be \textcolor{myp}{\textbf{ba717e}}.}
    \label{fig:input_data}
    % \vspace{-5mm}
\end{figure}

\paragraph{Model Architecture and Hyper-parameters.}
We train small Llama models~\cite{touvron2023llamaopenefficientfoundation} on a subset of the RedPajama dataset~\cite{together2023redpajama}, comprising 40 billion tokens, with the task of next-token prediction.
Our model has 162M parameters (12 layers, each with 12 attention heads; The hidden state dimension is 768, and the intermediate dimension of MLP layers is 3,072.)
The context length is 1,024 tokens. 
We use the Llama tokenizer with a vocabulary of 32,000 tokens.
Unless otherwise specified, the following hyperparameters are used:
The AdamW optimizer~\cite{adamw} with $(\beta_1, \beta_2)$ = (0.9, 0.999), a learning rate of 0.1, 2000 warmup steps, and the norm clip value of 1. 
Our training is conducted on 8 A100 GPUs, with a batch size of 64 per GPU.

\paragraph{Evaluating Context Copying.}

Each test sample consists of 50 random-token sequences, which are concatenated to form a single long sequence.
These sequences have an average length of 18 tokens, and we ensure that the 
$12$-gram prefix and $6$-gram suffix of each sequence is unique.
We append the prefix of the $i$-th sequence to the end of the concatenated sequences, which together serve as the model's input.
An example input case is shown in Figure~\ref{fig:input_data}.
Our test set includes 500 samples.

We ask the model to continue the input. 
An output is correct if it copies the suffix of the queried prefix from the context, since random token sequences lack meaningful semantics and the most natural continuation is to generate the suffix of the prefix that has appeared in the context~\cite{olsson2022context}.
To comprehensively assess context copying capabilities across different contextual positions, we evaluate the model for every $i$ mod $5 = 0$. 
Unless specifically indicated, we report the average accuracy across these positions, from models trained with 3 different random seeds.

\begin{figure}[t]
% \vspace{-9mm}
    \centering
    \includegraphics[width=\linewidth]{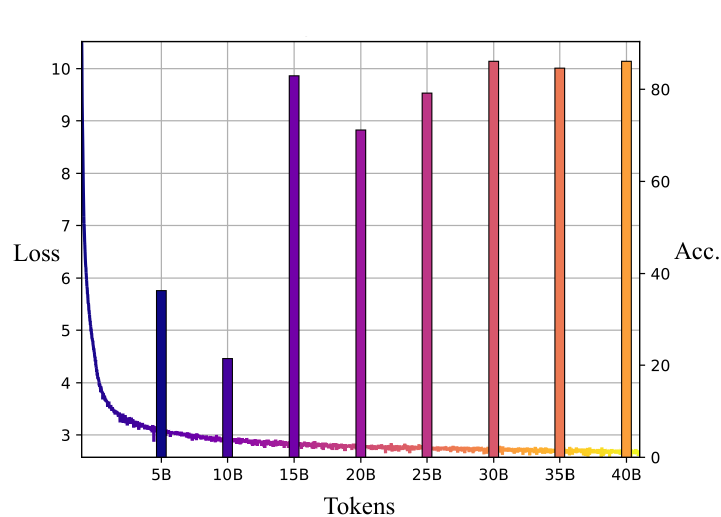}
    \caption{We illustrate the average context copying accuracy by the bars, and the pre-training loss by the line.
    The X-axis represents the number of tokens trained.
    A clear grokked copying occurs at 15B tokens.}
    \label{fig:acc_loss}
\end{figure}

\begin{figure}[t]
    \centering
    \includegraphics[width=\linewidth]{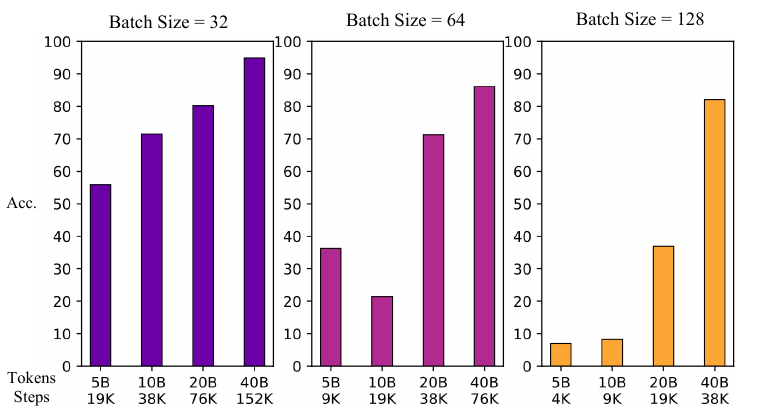}
    \caption{
    We manage the token count trained at specific steps by adjusting the batch size. 
    Three models trained with different batch size develop fundamental copying abilities after around 38,000 update steps, despite training on varying numbers of tokens.
    }
    \label{fig:acc_bsz}
    % \vspace{-5mm}
\end{figure}

\begin{figure}[t]
% \vspace{-9mm}
\centering
    \includegraphics[width=0.9\linewidth]{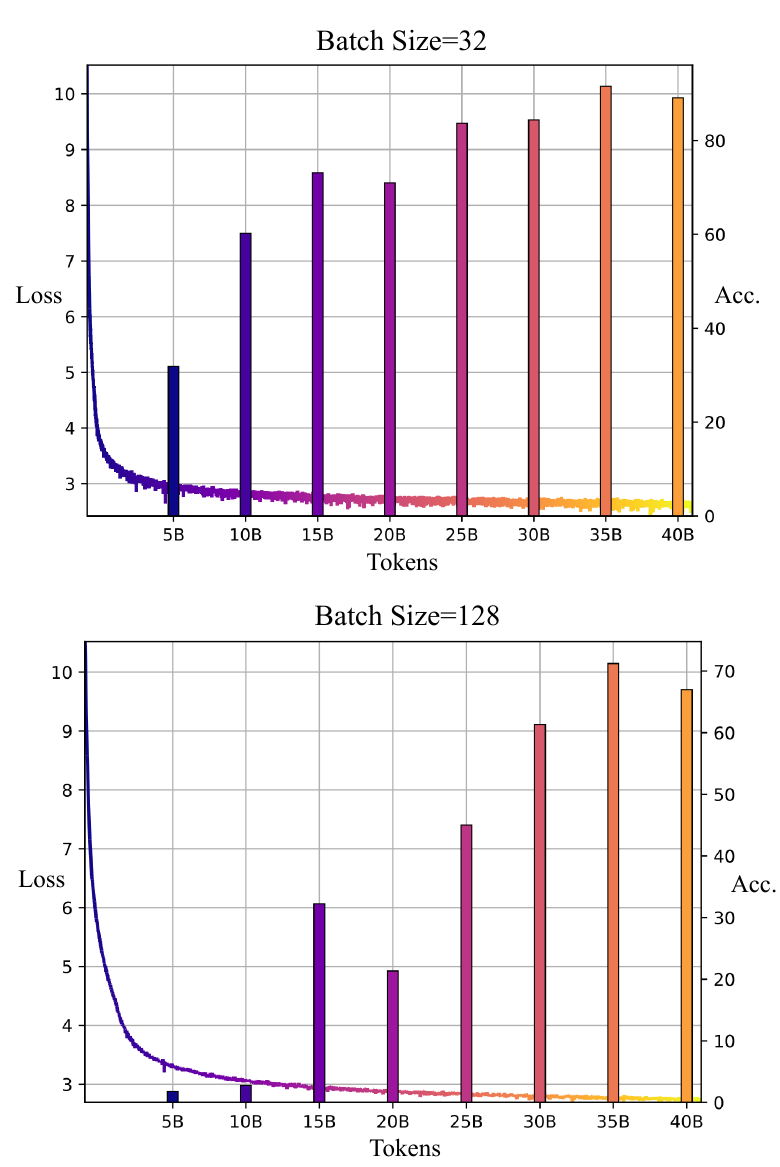}
    \caption{With a fixed learning rate,the convergence rate on the training set, as indicated by the training loss, is related to the token count.
    However, under similar convergence rates, the copying capacity varies significantly, which is influenced by the number of update steps.}
    \label{fig:acc_bsz_loss}
    % \vspace{-3mm}
\end{figure}

\section{Language Models ``Grok'' to Copy}
We propose that language models develop context copying in a manner similar to ``grokking''.
This section presents three arguments, along with supporting experiments and analyses.

\textbf{For Argument 1}, we present the context copying accuracy and pre-training loss in Figure~\ref{fig:acc_loss}.
The training loss stabilizes after 5B tokens, indicating that the fundamental language modeling has been established (i.e., fitted to the training distribution).
However, the accuracy is low until 10B tokens have been trained.
A surge in accuracy occurs at 15B tokens.
This pattern of developing robust context copying resembles grokking~\cite{power2022grokkinggeneralizationoverfittingsmall}.

\textbf{For argument 2}, we trained another two models using the same setups and same initial weights as described in Section~\ref{sec:setup}, but with batch sizes of 32 and 128. 
Our results indicate that grokked context copying is independent of the token count.
Figure~\ref{fig:acc_bsz} shows that \textit{with a fixed learning rate}, to achieve similar accuracy to models using a batch size of 64, models trained with a batch size of 128 (32) require twice (half) the token count, as their update steps are equal.
This finding aligns with observations~\cite{wang2024grokkedtransformersimplicitreasoners} that data quantity does not affect the grokking speed.
The consistency enhances the connection between grokking and the development of context copying.

Notably, we observed that the convergence on the training set is token-count-dependent, although copying performance is slowed down with larger batch sizes, as shown in Figure~\ref{fig:acc_bsz_loss}.
We assume that using an appropriately smaller batch size to update the models with more steps within a single epoch may facilitate the development of capacities that are not reflected in the training loss reduction.

\begin{figure}[t]
% \vspace{-8mm}
    \centering
    \includegraphics[width=0.9\linewidth]{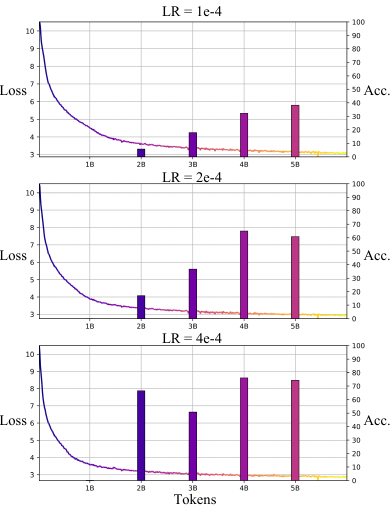}
    \caption{With a fixed batch size (64), a larger learning rate accelerates the grokking to copy.}
    \label{fig:lr}
    % \vspace{-4mm}
\end{figure}

Moreover, we examine the impact of learning rates. 
Figure~\ref{fig:lr} indicates that an increased learning rate facilitates earlier and stronger grokking.
Consequently, we assume that the grokked context copying doesn't emerge until the optimization reaches a specific intensity, which is influenced by both the learning rate and the number of update steps.

\textbf{For argument 3}, we examined the evolution of induction heads in our models.
Induction heads~\cite{elhage2021mathematical} are the primary circuit for conditional copying in Transformer-based language models and have been identified as a general mechanism across various models~\cite{lv2024interpretingkeymechanismsfactual}. 
Consider a sequence ``$A,B,...,A$'' input to the language model, where $A$ and $B$ are arbitrary tokens. 
Induction heads work based on collaboration across layers, enabling the model to output $B$. 
In shallower layers, certain attention heads move each token's information to its next position; in deeper layers, induction heads at the final position (i.e., the second $A$) attend to $B$ (since a subspace of hidden states at $B$'s position contains information from the first $A$) and copy the attended $B$ as the output. 

We introduce the induction score $I^{(L,H)}$, which quantifies the similarity between the behavior of the $H$-th head in layer $L$—referred to as $(L,H)$—and that of an ideal induction head. 
We establish $I^{(L,H)}$ as a value within the range of $[-1, 1]$, defined as:
\begin{align}
    I^{(L,H)} = \bar{A}^{(L,H)} \cdot EP^{(L,H)}.
    \label{eq:i}
\end{align}
In Eq.~\ref{eq:i}, \( \bar{A}^{(L,H)} \in [0, 1] \) measures the induction attention pattern: when inputting a random token sequence of length \( 2s \) which contains two identical subsequences of length $s$ (set to 100), we denote the average attention weight assigned from position \( s+i-1 \) to \( i \) as \( \bar{A}^{(L,H)} \), \( i \in [1, s-1] \).
Induction heads are expected to exhibit a high \( \bar{A}^{(L,H)} \) score.

$EP^{(L,H)} \in [-1, 1]$ in Eq.~\ref{eq:i} is the eigenvalue positivity of the OV circuit~\cite{elhage2021mathematical} of the head: $EP^{(L,H)} = \sum_i \lambda_i / \sum_i |\lambda_i|$.
$\lambda_i$ is the $i$-th eigenvalue of $(W_{U} W^{(L,H)}_{O} W^{(L,H)}_{V} W_{E})$, and $W^{(L,H)}_{O}$ and $W^{(L,H)}_{V}$ are weights of the value and output projection in head ($L,H$), while $W_{E}$ and $W_{U}$ are model's embedding and unembedding matrices.
A high $EP^{(L,H)}$ implies that the head copies the tokens it attends to as output.
Overall, a higher $I^{(L,H)}$ indicates a stronger induction head.

\begin{figure}[t]
% \vspace{-9mm}
    \centering
    \includegraphics[width=\linewidth]{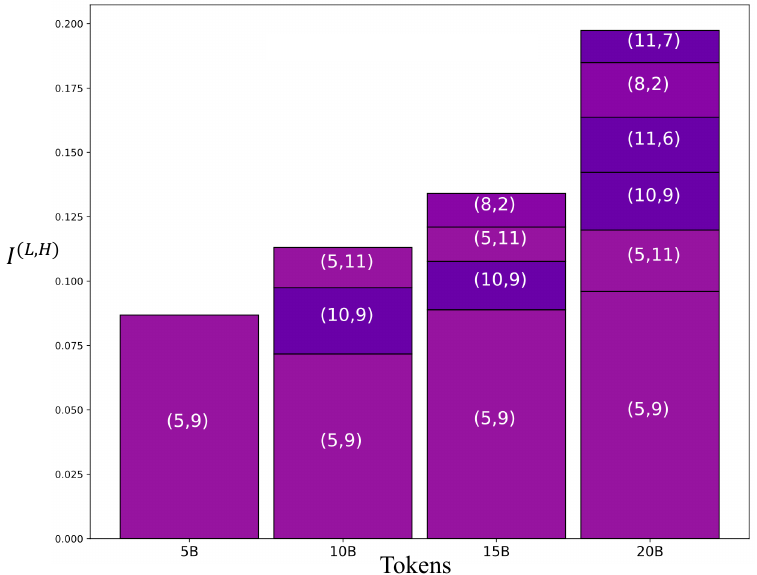}
    \caption{The evolution of induction heads during training. 
    A bar's height represents the $I^{(L,H)}$ value.
    Bars exhibiting larger values positioned nearer to the X-axis. 
    The results in this figure are from a single model.}
    \label{fig:head}
    % \vspace{-3mm}
\end{figure}

Figure~\ref{fig:head} illustrates the evolution of induction heads during training, revealing that they develop from shallower to deeper layers. 
This findings echos~\citet{wang2024grokkedtransformersimplicitreasoners}, who proposes that after grokking, models develop circuits in deeper layers.

\begin{figure}[t]
    \centering
    \includegraphics[width=\linewidth]{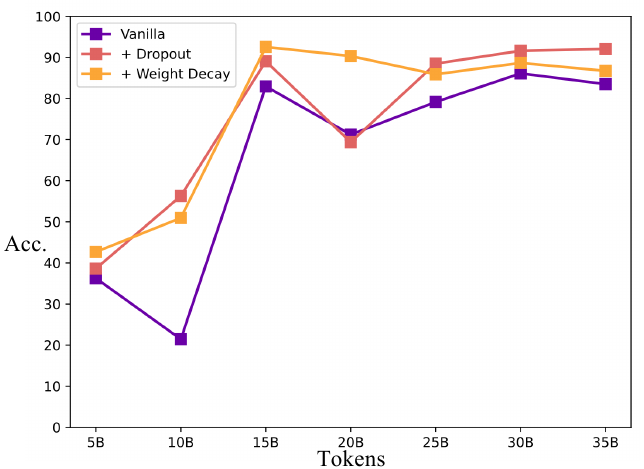}
    \caption{Regularization positively impacts the grokked copying.
    Compared with vanilla models, dropout accelerates the grokking process, advancing the abrupt accuracy increase from 15B tokens to 10B tokens, albeit with increased fluctuation in the evolutionary dynamics. 
    Both techniques improve the final accuracy.
    }
    \label{fig:reg}
    % \vspace{-4mm}
\end{figure}

\section{Application}
\label{sec:apply}

Viewing the development of context copying as a special grokking inspires us to examine the impact of regularization, as it enhances grokking~\cite{ nanda2023progressmeasuresgrokkingmechanistic}.
We train models using (1) 10\% attention dropout and (2) weight decay ($\lambda=0.1$). 
Figure~\ref{fig:reg} shows that their positive impact: with dropout, the model groks to copy earlier; both techniques improve the accuracy compared to the vanilla model. 

\section{Discussions}
We sincerely appreciate the anonymous reviewers for their valuable feedback. 
In this section, we address key points raised in their reviews, which may also be of interest to a broader audience.

\paragraph{1. Our motivation for using copying tasks to measure in-context ability.}

Induction heads, the key components responsible for in-context learning, are known to perform ``copy and paste,'' as described by~\cite{olsson2022context}. 
In essence, induction heads ``complete the pattern'' by copying and extending sequences that have occurred previously. 
This behavior motivates our exploration of copying, which are foundational to understanding in-context abilities.

Moreover, the copying task employed in this study has proven effective in previous research on RAG~\cite{tan2025pear} and in-context abilities~\cite{chen2024hopenovelpositionalencoding}.

\paragraph{2. We suggest evaluating grokking through downstream performance rather than training loss.}

In our task, the training objective is natural language modeling, while the testing task focuses on general copying. 
As a result, the training loss doesn’t fully capture the performance saturation seen in traditional grokking tasks. 
This is because copying can be viewed as a skill learned during pretraining, and once copying proficiency saturates, further improvements in other abilities can still lead to a decrease in training loss.

To demonstrate that copying on the training data has reached saturation, we measured the ``ICL score'' proposed by~\cite{olsson2022context}, which tracks the development of in-context abilities. 
Our results show that after approximately 4,000 training steps (about 1.85 billion tokens), the ICL score stabilizes at -0.5 nats. 
Since testing accuracy continues to improve well after this saturation point, we infer that once copying accuracy is ``grokked,'' reductions in training loss primarily stem from improvements in other abilities, rather than further progress in in-context copying.

\paragraph{3. The trade-offs between knowledge acquisition and in-context ability.}

Some studies~\cite{chang2024largelanguagemodelsacquire} suggest that large batch sizes enhance knowledge acquisition but hinder the development of in-context abilities, highlighting a trade-off between the two~\cite{nafar2024learningvsretrievalrole,yu-etal-2023-characterizing}.
While large batch sizes slow down in-context ability acquisition, their overall effect in real-world applications remains difficult to quantify, necessitating further research.

\paragraph{4. Properties of Grokking}
The properties of grokking are not limited to the three arguments we have exemplified.
Many studies~\cite{miller2024grokkingneuralnetworksempirical,fan2024deepgrokkingdeepneural,NEURIPS2022_dfc310e8,lee2024grokfastacceleratedgrokkingamplifying} explore various aspects of grokking; we list some for readers who may be interested.

\section{Conclusions}
This paper introduces a novel perspective that the development of context copying is a special grokking.
It holds the potential to provide meaningful insights that can be applied to language models, as we did in Section~\ref{sec:apply}.
We hope a better understanding of grokking in future works provide more insights for developing stronger language models.

\section*{Limitations}
This paper focuses on the copying task to reflect the development of in-context capacities.
Future innovations on improving the language model with better in-context capacities (e.g., ICL) might benefit from the correlations with grokking.
However, it is important to note that ICL presents a higher level of complexity compared to simple copying tasks. 
Due to our limited computational resources, we were unable to train language models to achieve robust ICL performance, and therefore did not evaluate ICL tasks.

\section*{Acknowledgement}

Ruobing Xie is supported by the Young Elite Scientists Sponsorship Program by CAST (2023QNRC001).
Ang Lv is supported by the Outstanding Innovative Talents Cultivation Funded Programs 2024 of Renmin University of China.

\bibliography{anthology,custom}

\end{document}